\pdfoutput=1
\documentclass[11pt]{article}

\usepackage[preprint]{acl}

\usepackage{times}
\usepackage{latexsym}

\usepackage[T1]{fontenc}
\usepackage[utf8]{inputenc}

\usepackage{microtype}

\usepackage{inconsolata}

\usepackage{graphicx}

\usepackage{amsmath}
\usepackage{amssymb}
\usepackage{tabularx}
\usepackage{multirow}

\title{Towards Time Series Generation\\ Conditioned on Unstructured Natural Language }

\author{Jaeyun Woo \and Jiseok Lee \and Brian Kenji Iwana \\
        Graduate School of Information Science and Electrical Engineering \\ 
        Kyushu University, Fukuoka, Japan \\ 
        \texttt{\{jaeyun.woo, jiseok.lee\}@human.ait.kyushu-u.ac.jp} \\
        \texttt{iwana@ait.kyushu-u.ac.jp}}

\begin{document}
\maketitle
\begin{abstract}
Generative Artificial Intelligence (AI) has rapidly become a powerful tool, capable of generating various types of data, such as images and text. However, despite the significant advancement of generative AI, time series generative AI remains underdeveloped, even though the application of time series is essential in finance, climate, and numerous fields. In this research, we propose a novel method of generating time series conditioned on unstructured natural language descriptions. We use a diffusion model combined with a language model to generate time series from the text. Through the proposed method, we demonstrate that time series generation based on natural language is possible. The proposed method can provide various applications such as custom forecasting, time series manipulation, data augmentation, and transfer learning. Furthermore, we construct and propose a new public dataset for time series generation, consisting of 63,010 time series-description pairs.

\end{abstract}

\section{Introduction}

Recently, generative artificial intelligence (AI) has advanced rapidly. 
For instance, Diffusion Models \citep{song2020denoising,yang2023diffusion} have recently become the base of generating images conditioned on natural language text.
Furthermore, methods such as Stable Diffusion \citep{rombach2022high} and DALL-E \citep{ramesh2021zero} have brought image generation into regular daily life.

However, compared to the image and text domains, the application of generative AI for time series remains underdeveloped.
Time series are patterns that are constructed from information in a sequence over time intervals. 
Time series are an important data type because they exist in many applications, such as signal processing, forecasting, bioinformatics, finance, gesture recognition, and more.

\begin{figure}%[h!]
    \centering
    \includegraphics[width=\columnwidth]{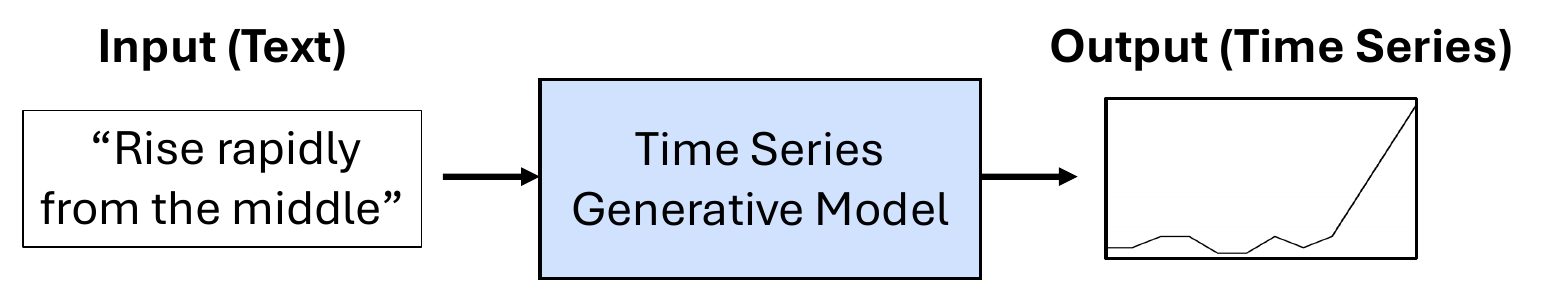}
    \caption{Time series generation conditioned on text.}
    \label{fig:goal}
\end{figure}

Despite their significance, time series generation is only a budding field. 
Currently, there are no generative models that can generate time series from unstructured natural language. 
Thus, as shown in Fig.~\ref{fig:goal}, a method of time series generation that is conditioned on text is proposed. 

Time series generated from natural language has many uses. 
For example, 
% as shown in Fig.~\ref{fig:U-Net Architecture}, 
it can be used for intelligent time series generation, custom forecasting, data augmentation, data manipulation, modality alignment, foundation models, and more.
In addition, unlike other time series generation models, the use of diffusion models allows for unconstrained, unstructured, natural language.

% \begin{figure}%[h!]
%     \centering
%     \includegraphics[width=0.6\columnwidth]{./images/motiv1.pdf}
%     \includegraphics[width=0.6\columnwidth]{./images/motiv2.pdf}
%     \caption{Motivations for generating time series.}
%     \label{fig:U-Net Architecture}
% \end{figure}

The proposed method utilizes a diffusion model with a language model to generate time series.
Diffusion models \citep{song2020denoising} are a type of generative model that synthesize data using a denoising process. 
The model operates in two phases.
First, there is a forward process, where Gaussian noise is incrementally added to an input pattern. 
Second, there is a reverse process, where a neural network learns to denoise and reconstruct the original data. 
Using this iterative denoising process, it is possible to generate high-quality samples.

% However, time series generation has unique challenges that image generation does not face. 
% Some of these challenges are common issues with time series generation, such as issues with variable lengths, temporal distortions, and time series structures.
% Other challenges are due to issues with diffusion models with time series, such as noise and denoising might not be appropriate for time series.

The contributions of this paper are as follows:
\begin{itemize}
    \item A temporal diffusion model is proposed. We demonstrate that the proposed model is successful at generating reasonable time series based on natural language descriptions.
    \item It is the first time that time series are generated from unconstrained and unstructured text. 
    \item We construct and propose a new public dataset for time series generation, consisting of 63,010 time series-description pairs, from a variety of sources including stock, synthetic data, signals, gestures, etc. The dataset and code can be found at https://github.com/[anonymous].
\end{itemize}

\section{Related Work}
\subsection{Time Series Generation}

\subsubsection{Statistical Models}
Time series generation has a rich history. 
Many of the early models utilize statistical, mathematical, or stochastic models to generate time series.
Some early works include posterior sampling \citep{Tanner_1987,Meng_1999} and linear models with probabilistic hyperparameters \citep{Fr_hwirth_Schnatter_1994}. 
More recent approaches include Local and Global Trend (LGT) \citep{smyl2016data}, which is a forecasting model that uses trends to model the time series, and GeneRAting TIme Series (GRATIS) \citep{Kang_2020}, which uses a mixture autoregressive (MAR) model to generate non-Gaussian and nonlinear time series.

\subsubsection{Generative Neural Networks}

There have been attempts to generate time series using Generative Adversarial Networks (GANs) \citep{goodfellow2020generative} and Variational Autoencoders~(VAE) \citep{kingma2013auto}. 
GANs are generative models that generate realistic data similar to the given dataset. 
% GAN consists of two neural networks, a generator and a discriminator. 
% The generator aims to produce synthetic data that is similar to the real data. The discriminator aims to distinguish the real and synthetic data. 
% To train the GAN, the generator and discriminator are trained in an adversarial manner.
% This interaction between two network makes it possible to generate super realistic synthetic data. 
% Alternatively, VAEs use an encoder-decoder structure. 
% VAEs are a subset of autoencoders that enforce a latent space with a normal distribution.
% Specifically, the encoder of a VAE encodes data into latent space, and the decoder decodes the latent space to generate time series.
% By using the decoder with sampled latent vectors, a new time series can be generated.
For an example using a GAN, \citet{yoon2019time} proposed TimeGAN.
TimeGAN uses a combination of a GAN and an autoencoder.
In the autoencoder, a time series is encoded and recovered. 
The TimeGAN then combines the autoencoder's latent space with the latent space of the generator. 
In this way, TimeGAN is specially designed to generate realistic time series data.
SeriesGAN \citep{eskandarinasab2024seriesgan} builds on TimeGAN and adds a second discriminator to improve the generated features.
GANs have also been used to generate time series in a number of domains, such as biosignals \citep{esteban2017real,Haradal_2018}, financial data \citep{wiese2020quant}, network traffic \citep{hasibi2019augmentation}, etc.

Controllable Time Series~(CTS) \citep{bao2024towards} use a novel VAE structure to create conditional time series. 
It is done by using a condition mapping in order to add controllable parameters to the VAE's latent space. \citet{desai2021timevae} use a multivariate VAE to generate time series.
Also, similar to GANs, VAEs have been used in a variety of applications, such as financial data \citep{acciaio2024time}, forecasting \citep{mentzelopoulos2024variational}, and gene modeling \citep{mitra2021rvagene}.

\subsection{Time Series and Language Models}

% -------------------
There have only been some works that use language models with time series, and most of these models do not model time series on language, like the proposed method.
Typically, the interaction between language models and time series is either not for generation, such as question-answering \citep{philpot2002dgrc,jin2020forecastqa} and captioning \citep{jhamtani2021truth}, or the language model is only used for pre-training, such as model alignment \citep{zheng2024revisited} or transfer learning \citep{garza2023timegpt}. 
Large Language Models~(LLM) can also be used for forecasting~\citep{gruver2023large,jin2023time}.
\citet{dong2024can} also tried to use LLMs for time series anomaly detection.

For diffusion models with language models, existing models are used for specific time series, such as text-to-audio \citep{lakhotia2021generative,richter2023speech,yang2024survey}. As far as the applicants know, this would be the first time that a generative model will be able to create time series based on unstructured natural language descriptions.

\section{Methodology}
\subsection{Diffusion Models}

\begin{figure*}%[h!]
    \centering
    \includegraphics[width=
    1.9\columnwidth]{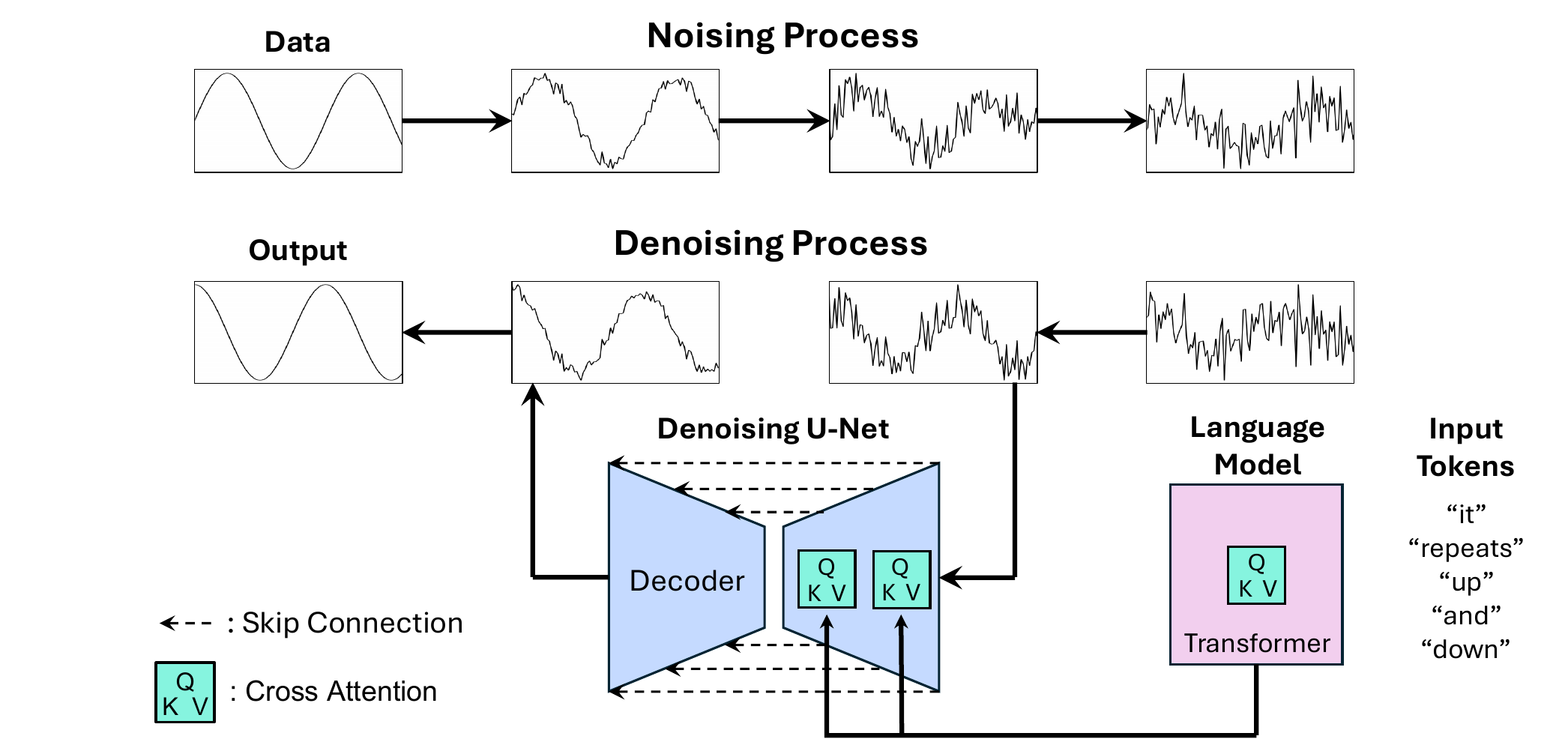}
    \caption{The proposed time series diffusion model}
    \label{fig:diffusion_model}
\end{figure*}

The diffusion model is a machine learning algorithm that generates data from the noise based on its pre-trained denosing process. As shown in Fig.~\ref{fig:diffusion_model}, it is trained in two phases. The first is the noising process that gradually adds Gaussian noise to the data.
% Gaussian noise is gradually added to the data. 
Namely, the original data \(x_0\) is converted into a noisy version \(x_t\) through Gaussian noise, or:
\begin{equation}
q(x_t \mid x_{t-1}) = \mathcal{N}(x_t; \sqrt{\alpha_t} x_{t-1}, (1-\alpha_t)\mathbf{I}),
\end{equation}
where $x_t$ is the noisy data sample at step $t$, $\mathcal{N}$ is Gaussian distribution, $\alpha_t$ is noise scaling parameter at step $t$ and $\mathbf{I}$ is the identity matrix.

Next is the denoising process. The model learns to denoise the data to reconstruct the original data, or:
\begin{equation}
p(x_{t-1} \mid x_t) = \mathcal{N}(x_{t-1}; \mu_\theta(x_t, t), \Sigma_\theta(x_t, t)),
\end{equation}
where $\mu_\theta$ is the learned mean parameter and $\Sigma_\theta$ is the learned covariance parameter.

The denoising process is done using a U-Net \citep{ronneberger2015u}, which consists of an encoder and a decoder structure.
In the encoder, the input data is progressively downsampled through a series of convolutional layers. 
At each downsampling step, important features are extracted and retained, capturing important characteristics of the input data.
In the decoder, the downsampled feature maps are gradually upsampled to reconstruct the original data. 
The encoder and decoder are connected by skip connections, which enable the decoder to access features that may have been lost during downsampling in the encoder. Skip connections ensure that important features are preserved, resulting in a more accurate and effective reconstruction during the upsampling process.

The model is trained by approximating the denoising process using a neural network. 
The network learns to predict the noise added at each step during the forward process by minimizing the mean squared error~(MSE) between the predicted and actual noise:
\begin{equation}
    L(\theta) = \mathbb{E}_{x_0, \epsilon, t} \left[ \| \epsilon - \epsilon_\theta(x_t, t) \|^2 \right],
\end{equation}
where $x_0$ is the original data, $x_t$ is the noised data at step $t$, $\epsilon$ is the noise added during the forward process and $\epsilon_\theta(x_t, t)$ is the neural network's predicted noise.
From this, the model can generate new data by applying this to the random noise.

\subsection{Language Conditioning}

% \subsubsection{Transformers}

% In this study, we trained the diffusion model with time series and its natural language description, and generated time series from it.
In order to condition the generated time series on natural language descriptions, a language model is used.
Similar to recent image-based diffusion models, the conditioning uses a transformer as the language model~\citep{waswani2017attention} with cross attention. 
As shown in Fig.~\ref{fig:diffusion_model}, cross attention is used between the transformer in the language model and the layers in the denoising U-Net.
Specifically, the attention of the final layer of the transformer is used with the output of a convolutional layer of the U-Net in: 
\begin{multline}
    \mathrm{Attention}(Q_\mathrm{LM}, K_\mathrm{DU}, V_\mathrm{DU}) \\
    =\mathrm{softmax}\left(\frac{Q_\mathrm{LM}K_\mathrm{DU}^T}{\sqrt{d_{k}}}\right)V_\mathrm{DU},
\end{multline}
where $Q_\mathrm{LM}$ (query), $K_\mathrm{DU}$ (key), and $V_\mathrm{DU}$ (value) are learned and weighted projections of the input of the layers, and $d_k$ is the dimensionality of $K_\mathrm{DU}$.
$Q_\mathrm{LM}$ is the input of the final layer of the language model transformer, $K_\mathrm{DU}$ and $V_\mathrm{DU}$ are feature maps taken from the Denoising U-Net. 
The output of this attention is then used for the next layer of the Denoising U-Net.

\subsection{Time Series with Diffusion Models}

We used a diffusion model to generate time series from unstructured natural language descriptions.
% The diffusion model is adapted to be used with time series.
This is done by modifying the Denoising U-Net to a temporal Denoising U-Net. 
Specifically, the 2D convolutions and 2D maxpooling layers meant for images are modified for time series by replacing them with temporal convolutions and 1D maxpooling layers, and the diffusion process and losses are adapted to be used with time series.

\section{Experimental Results}
\subsection{Datasets}

The proposed task is a new task. Thus, new datasets are required to be prepared in order to train the model.
Therefore, data was collected using two methods, a modification of the TRUth-Conditional gEneration~(TRUCE) \citep{jhamtani2021truth} dataset and a labeled dataset from gathered time series.

\subsubsection{TRUth-Conditional gEneration~(TRUCE)}
The TRUCE dataset, proposed by Jhamtani and Berg-Kirkpatrick \citep{jhamtani2021truth}, is a dataset that provides fixed-sized time series and natural language descriptions.
The intended task of the TRUCE dataset is time series captioning, i.e. generating text descriptions from a given time series.
This approach is the exact opposite of our research, as we are aiming to generate time series from natural language descriptions.
Thus, for our purposes, we reversed the usage of the data and labels, and used the text as the input and the time series as the target output. 

The TRUCE dataset includes 2,460 time series data with 12 time steps and three natural language descriptions for each. 
Using this, we can create a total of 7,380 pairs of time series and their descriptions. 
The dataset was split into time series independent training and test sets of 7,011 ($2337\times3$) and 369 ($123\times3$) time series, respectively.
In order to adapt to the proposed model, the time series are interpolated from 12 time steps to 100 time steps using linear interpolation.

\subsubsection{Labeled Dataset}

In addition to the TRUCE dataset, we gathered a larger dataset to increase the capability of the proposed method.
To label the gathered time series, a Generative Pretrained Transformer 4 omni~(GPT-4o) \citep{gpt4o} was used.
This is similar to the dataset augmentation method used by DALL-E 3~\cite{betker2023improving}.
First, time series were gathered from multiple sources. 
From each of these sources, time series with 100 time steps were gathered.
After the time series were gathered, the time series were rasterized as images for the GPT-4o model to describe. 

\begin{figure}[t]
 \begin{center}
    \includegraphics[width=1.01\columnwidth]{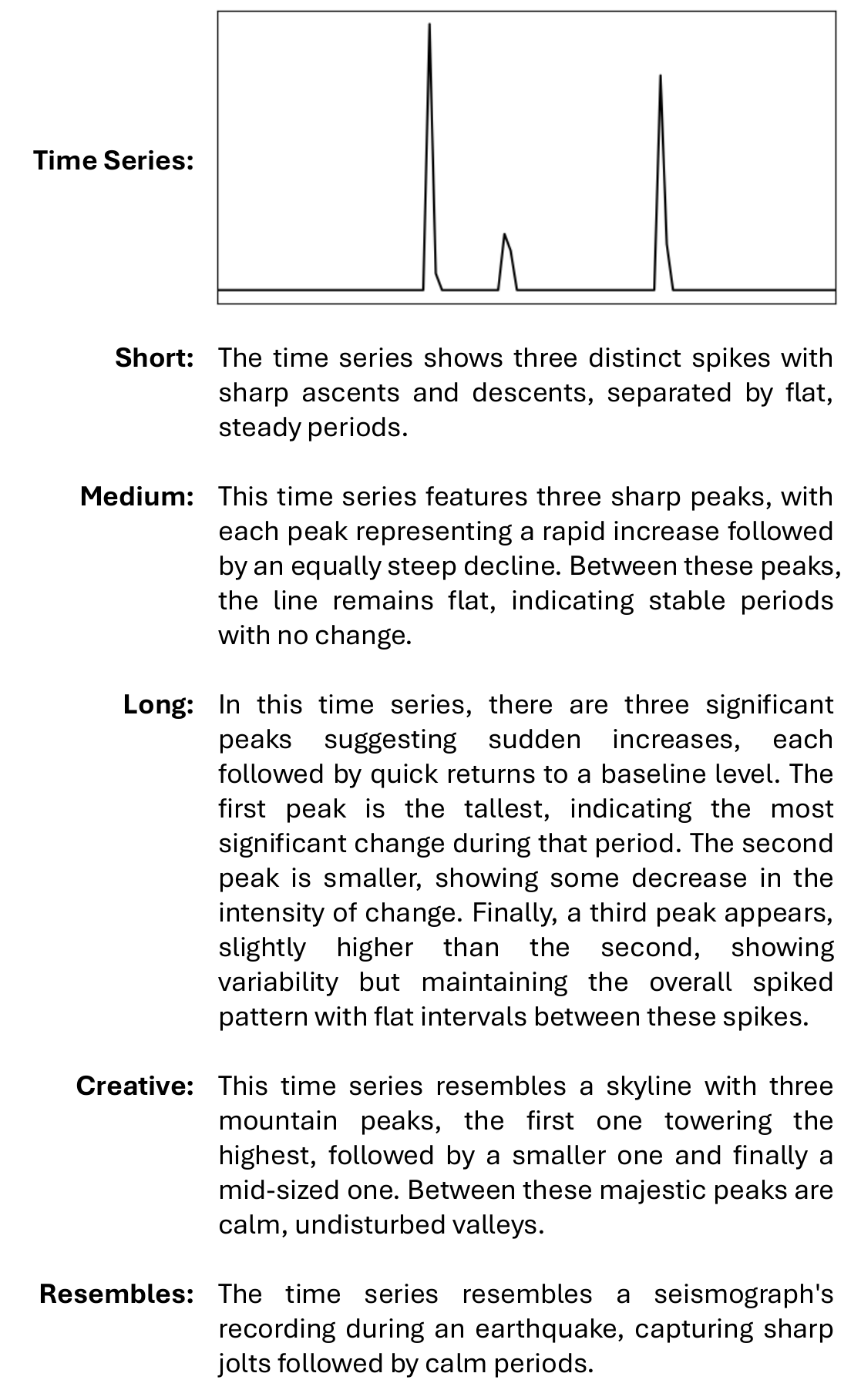}
    \caption{An example from the LargeKitchenAppliances dataset and the corresponding descriptions}
    \label{fig:examples_ucr}
 \end{center}
\end{figure}

Finally, the GPT-4o was prompted with the phrase, "Describe this time series with a short, medium, and long description. Make sure to describe the overall trends and changes in direction of the line. Also, a creative description and a description of what it resembles." and provided the rasterized time series and the expected format.
From this, we could generate five descriptions, a short, medium, long, creative, and resemblance descriptions for each time series.
Fig.~\ref{fig:examples_ucr} shows an example of the generated descriptions.
In total, a dataset of 63,010 time series-description pairs is created. 
It should be noted that the generation of the labels was performed in the image domain, but the dataset and proposed model only use the time series.

% \begin{figure}%[h]
%  \begin{center}
%     \includegraphics[width=1.0\columnwidth]{./images/stockexample.pdf}
%     \caption{A segment of the AAP stock and the corresponding descriptions}
%     \label{fig:examples_stock}
%  \end{center}
%   % \vspace{-2mm}
% \end{figure}

% The time series were gathered from Stock Market Data \citep{stockmarketdata}, the University of California Riverside (UCR) Time Series Classification Archive \citep{UCRArchive2018}. 
% In addition, a synthetic dataset was created by systematically adjusting the parameters of functions.

The data was collected from the following sources:

% \vspace{\baselineskip} 
\textbf{Stock Market Data.}
The Stock Market Data dataset \citep{stockmarketdata} contains stock market data.
% for companies listed on the National Association of Securities Dealers Automated Quotations~(NASDAQ) stock exchange, New York Stock Exchange~(NYSE), and Standard and Poor's 500 (S\&P500) stock trading index.
% The dataset contains the date, volume, high, low, and closing price for all of the listed companies in the exchanges or index at weekly intervals.
% 
For the data used in the experiment, only the closing prices of the companies in the New York Stock Exchange~(NYSE) are used. 
To preprocess the data, the closing price history of each company was cut into 100 time step increments. 
Any windows with gaps were excluded.
Also, due to the large number of gathered time series, only 1/10th, or 6,558 of the gathered time series, selected at random, are used.
The result was 6,558 1D time series with a variety of shapes for a total of 32,790 time series-description pairs. Among them, 31,150 pairs were used for training, and 1,640 pairs were used for evaluation.

% \vspace{\baselineskip} 
\textbf{UCR Archive Data.}
In order to add diversity to the data, we utilized University of California Riverside Time Series Classification Archive (UCR Archive) 2018 \citep{UCRArchive2018}.
The UCR Archive has 128 time series datasets across many domains, some of which include sensor data, electrocardiogram~(ECG), shape outlines, synthetic data, and more.
Due to the time series having varying lengths, all of the time series collected were resampled to 100 time steps regardless of the initial time series length.

For the data collection, 50 time series chosen at random from the training set of each dataset are used. 
For the datasets with 50 or fewer training set patterns, the entire training set was used. 
Only 50 or fewer time series from each dataset were used because many of the datasets have similar patterns, so this reduces redundancy as well as removes bias towards the larger datasets.

From the UCR Archive, 5,698 time series with 28,490 text descriptions were generated, split into 27,070 and 1,420 training and test sets, respectively.
% Figure~\ref{fig:examples_ucr} demonstrates examples of descriptions and time series.
It should be noted that the GPT-4o was not provided with any information about the type of time series or the source of the data. 
Any information in the descriptions about the type of data is inferred by the language model.

% \begin{figure*}%[h]
%  \begin{center}
%     \includegraphics[width=2.0\columnwidth]{./images/ucrexample.pdf}
%     \caption{An example from the ACSF1 dataset and the corresponding descriptions}
%     \label{fig:examples_ucr}
%  \end{center}
%   % \vspace{-2mm}
% \end{figure*}
% \vspace{\baselineskip} 
\textbf{Synthetic Data.}
To complement the data, we generated synthetic data using defined mathematical functions, such as linear, quadratic, cubic, and sinusoidal functions. By varying key parameters, we could create a diverse set of functions. 

For the linear equations, 100 time series are created by setting 100 discrete time steps $t$ sampled from:
\begin{equation}
    f(t)=m t,
\end{equation}
where $m$ is set to regular intervals between -1/50 and 1/50.
The interval ensures that the start and end of the time series lie in $f(t)\in[-1,1]$.
For the quadratic equations, 82 time series are created from:
\begin{equation}
    f(t)=m t^2,
\end{equation}
where the $m$ is selected from the 1st to 41st and 61st to 101st of 100 evenly spaced values between -1/2,500 and 1/2,500.
Similarly, the cubic equations are 82 time series sampled from:
\begin{equation}
    f(t)=m t^3,
\end{equation}
where $m$ is selected from the 1st to 41st and 61st to 101st of 100 evenly spaced values between -1/12,500 and 1/12,500.
Finally, the sinusoidal are generated from:
\begin{equation}
    f(t)=\mathrm{sin}(mt),
\end{equation}  11.17   13.3
where $m$ is selected from the 1st to 41st and 61st to 101st of 100 evenly spaced values between $-\pi$ and $\pi$.
For all functions, $m=0$ is avoided. As a result, a total of 346 time series were generated. 
% With GPT-4o model, we generated five descriptions for each time series. 
% Thus, 
1,730 synthetic time series descriptions were created, with 1,645 training and 85 test.

% \begin{figure}%[h]
%  \begin{center}
%     \includegraphics[width=1.0\columnwidth]{./images/syntheticexamples.pdf}
%     \caption{$f(t)=\mathrm{sin}\left(\frac{61}{50\pi}t\right)$ and the corresponding descriptions}
%     \label{fig:examples_synthetic}
%  \end{center}
%   % \vspace{-2mm}
% \end{figure}

\subsection{Architecture and Settings}

We trained the diffusion model with time series data and their natural language descriptions. 
The diffusion model was trained with 500 diffusion steps for 30 epochs. 
The model was trained using an Adam optimizer \citep{kingma2014adam} with an initial learning rate of 0.0001 and batch size of 64.
Training takes 1.5 hours on an NVIDIA TITAN RTX. Generation only takes a few seconds.

The Denoising U-Net used by the proposed method contains four convolutional layers in the encoding path and four transposed convolutional layers in the decoding path. 
The first convolutional layer has a filter size 3 and stride 1.
The subsequent convolutional layers are stride 2 in order to downsample the time series.
The transposed convolutional layers are the reverse of the convolutional layers.
Furthermore, skip connections are used between corresponding convolutional layers and transposed convolutional layers.
The Sigmoid Linear Unit~(SiLU) \citep{hendrycks2016gaussian} is used for the activation function on the convolutional layers, and transposed convolutional layers and GroupNorm \citep{wu2018group} are used between layers.

For the language model, a pretrained Bidirectional Encoder Representations from Transformers~(BERT) \citep{devlin2018bert} is used. 
The BERT model uses a vocabulary of 30,522 word piece tokens.
In order to facilitate the condition, cross-attention from the language model is used after the third and fourth convolutional layers and the first and second transposed convolutional layers.

\begin{figure}
 \begin{center}
    \includegraphics[width=1\columnwidth]{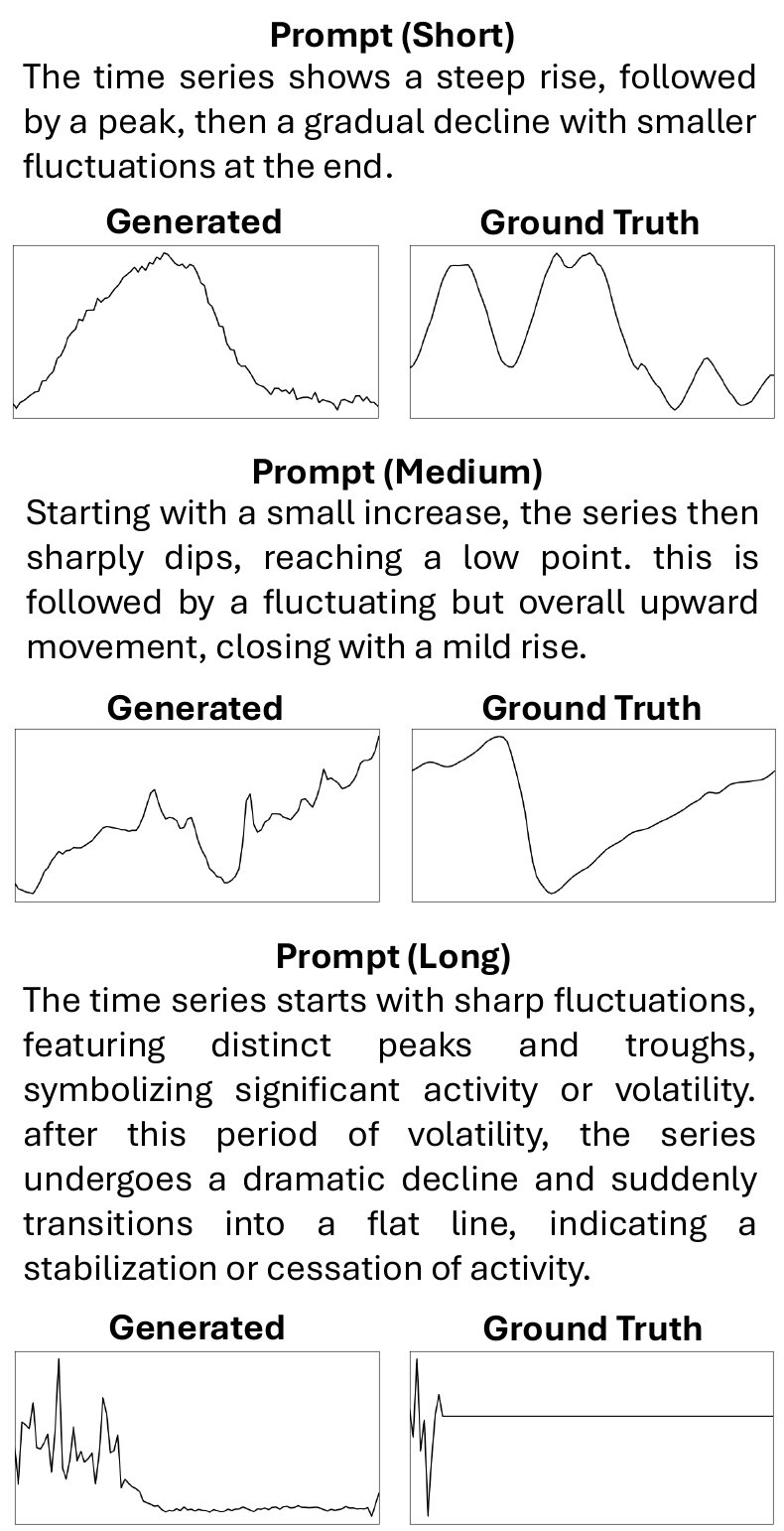}
    \caption{Examples of generated time series from Short, Medium, Long prompt from the test set.}
    \label{fig:smlresultsall}
 \end{center}
\end{figure}

% \subsection{Evaluation}

% We wanted to figure out how creative and resemble descriptions affects the training output, we conducted quantitative evaluation. 
% To be specific, we trained the model in 4 ways. With all descriptions, without Creative, without Resemble, and without creative and resemble descriptions.

\subsection{Results}
Fig.~\ref{fig:smlresultsall} shows example results of Short, Medium, and Long descriptions with their ground truth.
As shown in the figure, the proposed method was able to generate time series that matched the prompt. 
For example, in the Short prompt, the generated time series "shows a steep rise, followed by a peak, then a gradual decline." 
In this way, the generated time series was better than the ground truth.
The Medium and Long descriptions also closely followed the descriptions.

% In particular, generated time series of the long prompt followed well. 
% The long prompt asks for "an overall positive trend," which the time series has. 
% In addition, as the prompt asks, the time series "begins with an initial decline, followed by a sharp upward spike," and "post the rise, the data continues with moderate fluctuations comprising various peaks and valleys."

%%%%%%%%%%%%%%%%% Caution: Bad English %%%%%%%%%%%%%%%%%%%%%%%

\begin{figure}%[h]
 \begin{center}
    \includegraphics[width=1\columnwidth]{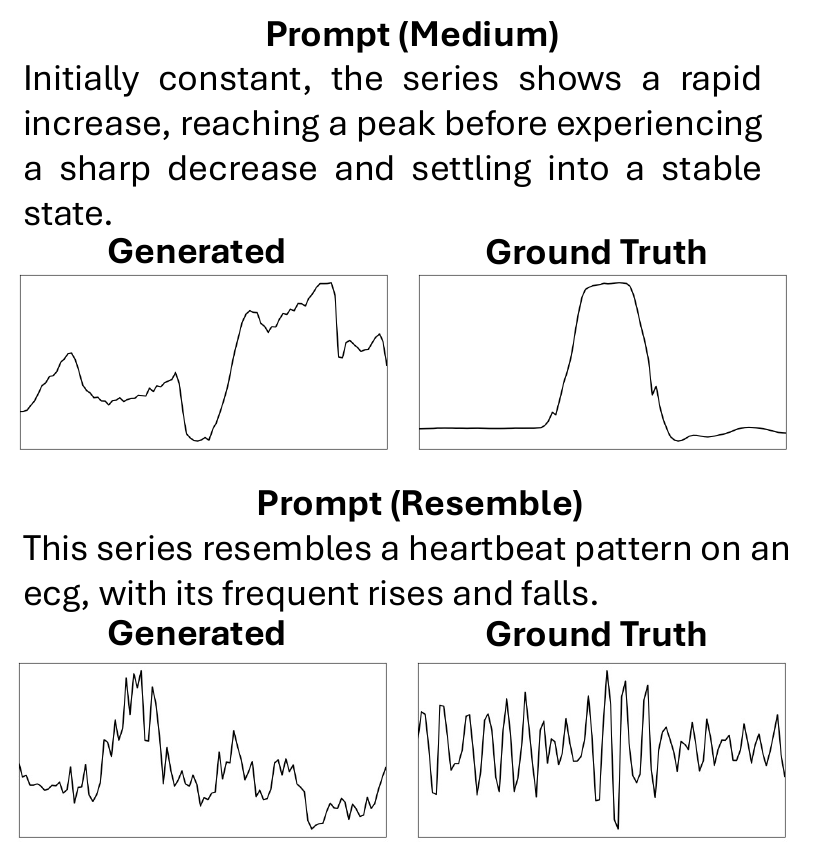}
    \caption{Example of generated time series with their prompts does not match the generated time series.}
    \label{fig:poorresults}
 \end{center}
  % \vspace{-2mm}
\end{figure}

However, not all of the results were successful. 
Fig.~\ref{fig:poorresults} shows some examples of poor results from a medium-length prompt and a Resemble prompt. In the Medium prompt, it asks for the time series to "experience a sharp decrease and settle into a stable state." However, the generated time series does not settle into a stable state, it keeps fluctuating after the sharp decrease.
Also, the resemble prompt asks for the time series to "resemble a heartbeat pattern", but the generated output does not.

%%%%%%%%%%%%%%%%%%%%%% bad English %%%%%%%%%%%%%%%%%%%%%%%%%%%

% In the figure, it can be seen that the generated time series matches the prompt well.
% In the first prompt, "the time series shows small fluctuations followed by a sudden spike and subsequent sharp decline, ending in slight ups and downs," the generated time series has many sharp spikes instead of just one like the prompt suggests. 
% However, the ground truth also lacks the sharp decline described.
% In the other examples, the prompt does not have enough information. 
% Either the descriptions are wrong, such as the middle example, or the description is too short to accurately portray the ground truth, such as the bottom example.
% This further establishes that unsurprisingly, the longer descriptions generate more accurate time series.

% \subsection{My own text generation}

\begin{figure}[t]
 \begin{center}
    \includegraphics[width=1\columnwidth]{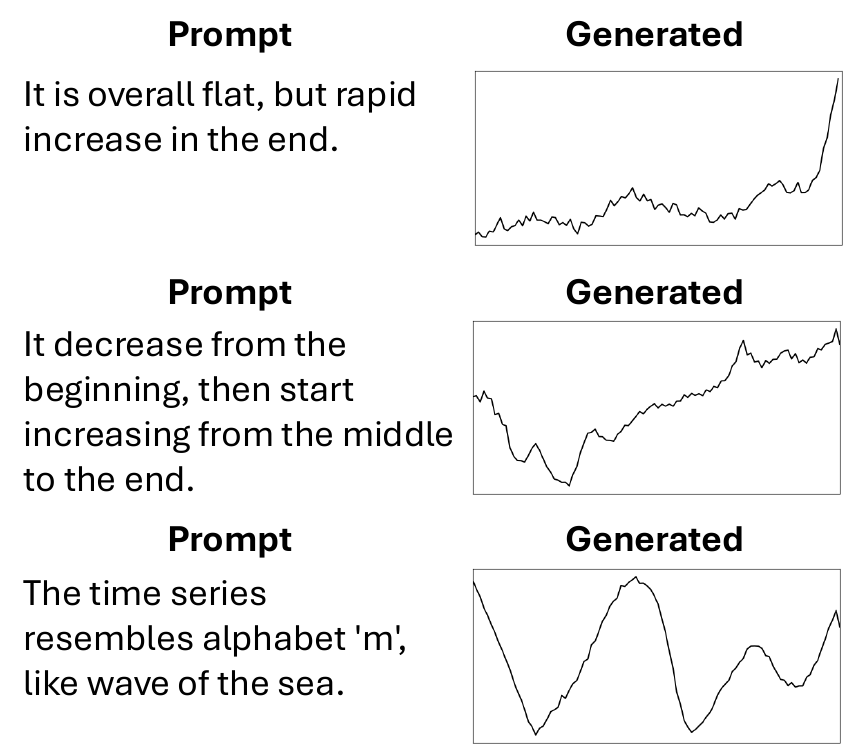}
    \caption{Examples of generated time series from custom text.}
    \label{fig:nobert}
 \end{center}
\end{figure}

\begin{figure}
 \begin{center}
  \includegraphics[width=1\columnwidth]{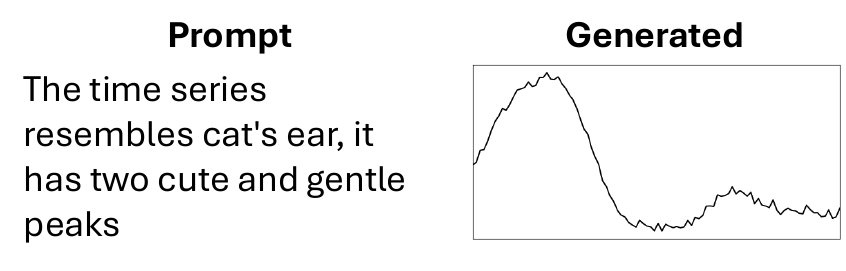}
  \caption{Example of a poor custom prompt.}
  \label{fig:cat}
  \end{center}
\end{figure}

In addition to prompts from the dataset, it is possible to provide the model with custom prompts. Example results are shown in Fig.~\ref{fig:nobert}.
% I tried to generate with my own text. 
% I felt like when generating from the words that were likely in the dataset, it performed well.
As instructed, the upper time series is "overall flat" but has a "rapid increase in the end," the middle "decreases from the beginning" and "start increasing from the middle."
Surprisingly, the bottom time series prompt successfully worked, despite the provided prompt describing a time series in a non-conventional way.
However, there are limitations to what can be input. 
For example, in Fig.~\ref{fig:cat}, the prompt was too far outside what could be found in the dataset. Thus, this is a limitation of the proposed method and will be discussed in the Limitations section.

\section{Discussion}

\subsection{Creative and Resembles Descriptions}

In addition to the different lengths of descriptions, we used two additional description types, a creative description and a description of what the time series resembles.
These descriptions were added to expand the capability of the proposed method to model abstract descriptions of what a time series might generate.

\begin{figure}[t]
 \begin{center}
    \includegraphics[width=1\columnwidth]{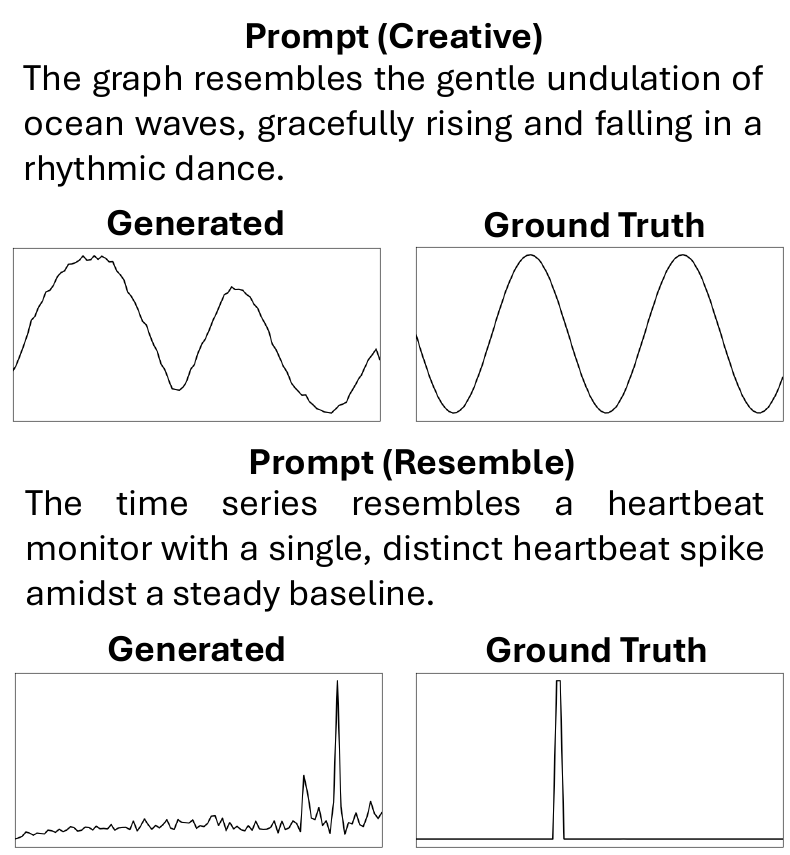}
    \caption{Examples of generated time series from Creative and Resemble prompts from the test set.}
    \label{fig:CRresultsall}
 \end{center}
  % \vspace{-2mm}
\end{figure}

As shown in Fig.~\ref{fig:CRresultsall}, the model was successful in generating time series that match the Creative and Resemble prompts. 
For example, the Creative prompt asks for "gentle undulation of ocean waves, gracefully rising and falling in a rhythmic dance" and the generated time series closely matches the ground truth. Also, the Resemble prompt asks for "heartbeat monitor with a single, distinct heartbeat spike amidst a steady baseline" and the generated time series is flat with one significant peak. 

However, to see whether adding the Creative and Reassemble descriptions affects the overall performance, we conducted a quantitative evaluation.
For the evaluation, we use Euclidean distance \citep{danielsson1980euclidean} and Dynamic Time Warping (DTW) distance \citep{sakoe2003dynamic} between the generated time series and the ground truth time series.
In this case, Euclidean distance is the pair-wise element distance between the time series, using the L1 distance between elements. 
DTW is similar to Euclidean distance except that it uses an elastic matching between the elements. This allows for the features to be optimally warped to match, i.e. the time series are similar if the main features are similar but not necessarily in the same place.
For DTW, we use an asymmetric slope constraint.

The results are shown in Table~\ref{dtw}.
It can be seen that for most of the subsets, including using the entire test dataset, the best model is trained using all of the training data. 
Although the model without training with Creative and Resembles data had lower Euclidean distances for the length-based prompts. 
This is reasonable due to the trained model without Creative and Resembles would be better for more literal descriptions. 
However, we believe that sacrificing slightly on the length-based descriptions is worth the model being able to handle more unconventional prompts.
% Table shows output of DTW and Euclidean Distance.
% According to the DTW, the model trained with all dataset performed best.
% However, according to the Euclidean distance, the model trained without creative and resemble descriptions performed the best. However, we can find out that although NO CR is performing best with Length-based model(Short, Medium, Long) It is most poor with Creative and Resemble prompts. Beside, All is performing stable in every fields. So, we may say that ALL is the best.

\begin{table*}%[ht]
  \centering
  \begin{tabular}{lc|c|cccc}
    \hline
     && & &  & {w/o Creative} &  \\
     && {Proposed} & {w/o Creative} & {w/o Resembles} & {\quad\& Resembles} & w/o TRUCE \\
    \hline
    \multirow{2}{*}{Short}
    & ED & 30.19 &30.97 &29.86 &\textbf{29.08} &31.09\\
    & DTW & 14.18	&14.54 &14.41 &14.16 &\textbf{13.78} \\
    \hline
    \multirow{2}{*}{Medium}
    &ED& 30.35& 30.41 &30.17	&\textbf{29.55} &31.46\\
    &DTW& \textbf{14.06}& 14.28 &14.68	&14.47 &14.17\\
    \hline
    \multirow{2}{*}{Long}
    &ED& 29.47& 30.77& 30.03& \textbf{29.03} &31.48\\
    &DTW& \textbf{13.59}& 14.53& 14.57& 14.07 &13.94\\
    \hline
    \multirow{2}{*}{Creative}
    &ED& 30.50&	\textbf{30.43}&	31.23&	31.64&	31.82\\
    &DTW& \textbf{14.45}&	14.72&	15.49&	16.63&	14.55\\
    \hline
    \multirow{2}{*}{Resembles}
    &ED&\textbf{30.34}&	31.01&	31.70&	32.70&	31.48\\
    &DTW&\textbf{14.22}&	14.81&	15.37&	17.06&	14.24\\
    \hline
    \multirow{2}{*}{TRUCE}
    &ED&32.53 &33.43 &33.27 &\textbf{32.07} &35.20\\
    &DTW&\textbf{15.33} &15.80 &15.46 &15.73 &16.21\\
    \hline
    \multirow{2}{*}{All}
    &ED & \textbf{30.76}&	31.39& 31.26 &30.82	&32.39\\
    &DTW & \textbf{14.41}& 14.88& 15.04 &15.39	&14.65\\
    \hline
  \end{tabular}
  \caption{\label{dtw}
    Distance between the generated time series and the ground truth based on the description type. The rows refer to subsets of the test data. The columns refer to models trained without Creative, without Resembles, without both, and without TRUCE.
  }
\end{table*}

\subsection{Quality of the Descriptions}
% \subsubsection{Length-based Descriptions}

\begin{figure}[t]
 \begin{center}
    \includegraphics[width=0.95\columnwidth]{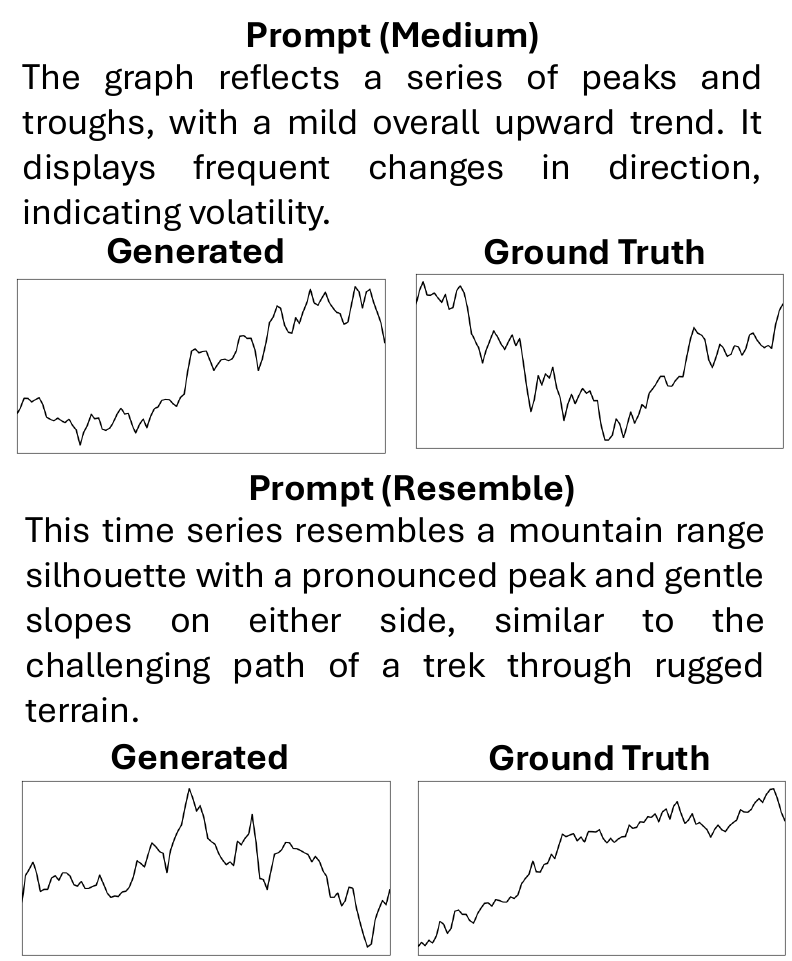}
    \caption{Examples of poor descriptions from the Labeled Dataset.}
    \label{fig:badprompts}
 \end{center}
  % \vspace{-2mm}
\end{figure}

% As noted previously,
The quality of the generated time series depends on the robustness of the data. 
However, some of the ground truth descriptions are poor.
The most notable problem is that some of the descriptions generated by GPT-4o are too generic or sometimes misleading. 
For example, in Fig.~\ref{fig:badprompts}, the upper prompt is more similar to the proposed method's generated time series than the ground truth. 
In the lower example, the prompt is ambiguous and could be used for many different time series.
In order to overcome this issue, future work will be done in cleaning the dataset.

% They are too generic like upper one, and both of them are even wrong. Even the generated time series is better than ground truth.

\section{Conclusion}
In this paper, we attempted to generate time series based on unconstrained and unstructured natural language text. 
To do this, we proposed to use a diffusion model utilizing a temporal denoising U-Net. 
The denoising U-Net is conditioned using a pretrained BERT as the language model.
Through the experiments, we demonstrated that well-formed time series can be generated successfully.

In addition, we created a dataset that contains 63,010 time series-description pairs. 
This dataset contains 32,790 pairs derived from pieces of stock tickers, 28,490 pairs from time series in the UCR Time Series Classification Archive, and 1,730 pairs from synthetic time series based on simple equations.
In the dataset, each pair consists of 100 time step time series and its natural language description.
The dataset is publicly available at https://github.com/uchidalab/ts-generation.

In the future, we plan on improving the proposed model. 
There are still outstanding issues that need to be addressed.
For example, the model lacks considerations for challenges that time series might encounter.
Another issue is that the text descriptions of the time series can be improved upon. 
This can be done with alignment or by manually editing the labels.

\newpage

\section*{Limitations}

As mentioned previously, there are some limitations on the prompts that can be used to produce high-quality results. 
The ability of the proposed model can be improved by adding more diverse descriptions.
However, there is a limit to what can be conceived.
In addition, while the proposed method uses time series from a variety of sources, such as the 128 datasets of the UCR Archive, synthetic data, and stock data, more types of time series can be added. 

A limitation of the proposed dataset is that it has many ambiguous descriptions.
To overcome this, in the future, we will perform a careful curation of the dataset.

Finally, the type of time series generated has some limitations. 
While the proposed method can be used for time series of any length or number of dimensions, there are still limitations that are common to time series, such as generating time series of different lengths from the same model.

% \section*{Acknowledgments}

% This document has been adapted
% by Steven Bethard, Ryan Cotterell and Rui Yan
% from the instructions for earlier ACL and NAACL proceedings, including those for
% ACL 2019 by Douwe Kiela and Ivan Vuli\'{c},
% NAACL 2019 by Stephanie Lukin and Alla Roskovskaya,
% ACL 2018 by Shay Cohen, Kevin Gimpel, and Wei Lu,
% NAACL 2018 by Margaret Mitchell and Stephanie Lukin,
% Bib\TeX{} suggestions for (NA)ACL 2017/2018 from Jason Eisner,
% ACL 2017 by Dan Gildea and Min-Yen Kan,
% NAACL 2017 by Margaret Mitchell,
% ACL 2012 by Maggie Li and Michael White,
% ACL 2010 by Jing-Shin Chang and Philipp Koehn,
% ACL 2008 by Johanna D. Moore, Simone Teufel, James Allan, and Sadaoki Furui,
% ACL 2005 by Hwee Tou Ng and Kemal Oflazer,
% ACL 2002 by Eugene Charniak and Dekang Lin,
% and earlier ACL and EACL formats written by several people, including
% John Chen, Henry S. Thompson and Donald Walker.
% Additional elements were taken from the formatting instructions of the \emph{International Joint Conference on Artificial Intelligence} and the \emph{Conference on Computer Vision and Pattern Recognition}.

\bibliography{custom}

% \appendix

% \section{Example Appendix}
% \label{sec:appendix}

% This is an appendix.

\end{document}